\documentclass[conference]{IEEEtran}

\IEEEoverridecommandlockouts
\usepackage{cite}
\usepackage{amsmath,amssymb,amsfonts}
\usepackage{graphicx}
\usepackage{textcomp}
\usepackage{xcolor}
\usepackage{subfigure}
\usepackage{enumerate}
\usepackage[flushleft]{threeparttable}
\usepackage{algorithm}
\usepackage{makecell}
\usepackage{float}
\usepackage[noend]{algpseudocode}
\usepackage{diagbox}
\usepackage{booktabs}
\usepackage{adjustbox}
\usepackage{comment}

\usepackage{tabularx}
\usepackage{hyperref}

\makeatletter 
\newcommand{\linebreakand}{%
  \end{@IEEEauthorhalign}
  \hfill\mbox{}\par
  \mbox{}\hfill\begin{@IEEEauthorhalign}
}
\makeatother 

\def\BibTeX{{\rm B\kern-.05em{\sc i\kern-.025em b}\kern-.08em
    T\kern-.1667em\lower.7ex\hbox{E}\kern-.125emX}}
\begin{document}



\title{A Survey on Data Quality Dimensions and Tools for Machine Learning \\ {\large Invited Paper} }


\author{\IEEEauthorblockN{1\textsuperscript{st} Yuhan Zhou}
\IEEEauthorblockA{\textit{Dept. of Information Science} \\
\textit{University of North Texas}\\
Denton, Texas, USA \\
yuhanzhou@my.unt.edu}
\and
\IEEEauthorblockN{2\textsuperscript{nd} Fengjiao Tu}
\IEEEauthorblockA{\textit{Dept. of Information Science} \\
\textit{University of North Texas}\\
Denton, Texas, USA \\
fengjiaotu@my.unt.edu}
\and
\IEEEauthorblockN{3\textsuperscript{rd} Kewei Sha}
\IEEEauthorblockA{\textit{Dept. of Information Science} \\
\textit{University of North Texas}\\
Denton, Texas, USA \\
kewei.sha@unt.edu}
\and

\linebreakand 

\IEEEauthorblockN{4\textsuperscript{th} Junhua Ding}
\IEEEauthorblockA{\textit{Dept. of Information Science} \\
\textit{University of North Texas}\\
Denton, Texas, USA \\
junhua.ding@unt.edu}
\and

\IEEEauthorblockN{5\textsuperscript{th} Haihua Chen \thanks{Corresponding author: Haihua Chen (haihua.chen@unt.edu). This research is partially supported by the College of Information’s Research Seed Grant in 2024 at University of North Texas.}}
\IEEEauthorblockA{\textit{Dept. of Information Science} \\
\textit{University of North Texas}\\
Denton, Texas, USA \\
haihua.chen@unt.edu}
}

\maketitle


\begin{abstract}
Machine learning (ML) technologies have become substantial in practically all aspects of our society, and data quality (DQ) is critical for the performance, fairness, robustness, safety, and scalability of ML models. With the large and complex data in data-centric AI, traditional methods like exploratory data analysis (EDA) and cross-validation (CV) face challenges, highlighting the importance of mastering DQ tools. In this survey, we review 17 DQ evaluation and improvement tools in the last 5 years. By introducing the DQ dimensions, metrics, and main functions embedded in these tools, we compare their strengths and limitations and propose a roadmap for developing open-source DQ tools for ML. Based on the discussions on the challenges and emerging trends, we further highlight the potential applications of large language models (LLMs) and generative AI in DQ evaluation and improvement for ML. We believe this comprehensive survey can enhance understanding of DQ in ML and could drive progress in data-centric AI. A complete list of the literature investigated in this survey is available on GitHub at: \url{https://github.com/haihua0913/awesome-dq4ml}. 
\end{abstract}


\section{Introduction}



The quality of the training data used in ML models has a significant impact on the performance, fairness, robustness, safety, and scalability of the models \cite{chenDataCuration2022,FENZA2021107366, RangineniAnalysis2023, GeigerGarbageInGarbageOut2020}.  Existing studies show that there is a causal relationship between improving the quality of the training data and increasing the performance of computer vision, classification, and other ML tasks \cite{rao2020quality, ChenEvaluation2021, budachEffectsDataQuality2022a}. Improving the quality of the dataset used in ML can be more efficient than enlarging the quantity \cite{joshi2024dataefficient, gudivadaDataQualityConsiderations}. Inaccurate, biased, and incomplete datasets may cause unfair, bad-performance ML models that cannot be used in decision-making and mislead downstream artificial intelligence (AI) applications \cite{budachEffectsDataQuality2022a, GuptaDataQuality2021}.


Data-centric AI values data maintenance, including data understanding, DQ evaluation and improvement, and storage and retrieval \cite{zha2023datacentric}. With this trend, traditional model work and general DQ metrics cannot satisfy the needs of DQ in ML anymore.  DQ tools can facilitate this progress through data profiling, DQ issue detection, and DQ monitoring. Therefore, it is necessary to investigate current DQ tools, analyze their strengths and limitations based on DQ metrics in ML, and further design useful frameworks.

Motivated by this, we investigate how DQ evaluation and improvement work in current DQ tools, which have been upgraded from the versions in \cite{ehrlingerSurveyDataQuality2022}. We also introduce a roadmap to design tools for DQ evaluation and improvement in ML. Overall, this survey covers both theoretical and empirical discussions with real-world applications, combined with emerging trends, such as LLMs and generative AI. The contributions of this paper are as follows: 

\begin{itemize}
    \item Presenting an overview of 4 DQ dimensions, and 12 metrics in ML, with the definitions, and examples.
    \item Reviewing 17 data quality evaluation and improvement tools in the last 5 years.
    \item Putting forward a development roadmap for the framework and function designs through a summary and comparative analysis of the tools.
\end{itemize}

The paper is organized as follows. Section \ref{sec:2} summarizes the definitions and impacts of DQ in ML and the dimensions used in the evaluation. In section \ref{sec:3}, we survey existing open-source DQ evaluation and improvement tools. From the comparative analysis, we further illustrate the development roadmap and core functions of designing new tools in section \ref{sec:4}, discussing the emerging trends and advancements.

\section{Related Work}


\begin{table*}
\scriptsize
\caption{\centering Existing surveys related
to data quality in machine learning}
\label{table:survey}
\setlength{\tabcolsep}{3pt}
\renewcommand\arraystretch{1.1}

\begin{threeparttable}
\begin{tabularx}{\textwidth}{p{0.8cm}p{0.6cm}p{3.9cm}X}
\toprule
   \textbf{Survey} & \textbf{Year} & \textbf{Topics} & \textbf{Findings}\\ 
   \midrule

   \cite{mohammed2024data} & 2024 & DQ assessment, challenges, and opportunities & Propose a comprehensive framework for systematically assessing data quality across 29 dimensions and 5 facets (the data itself, the data source, the system to access it, the task, and the interacting humans), and challenges including law and social sciences. \\

   \cite{LiActiveLearning2024} & 2024 & Active learning, data quality control, anomaly detection & Highlight that active learning (AL) offers a promising solution to data quality control (DQC) in machine learning (ML) by reducing the labeling burden on domain experts while presenting a review of common data quality issues and assessment methods, particularly in anomaly detection tasks. \\

   \cite{lu2023mlSynthetic} & 2023 & ML models for synthetic data generation & Present a systematic review of studies using machine learning for synthetic data generation, highlighting its applications across domains including education and health, and various data types. \\
 
   \cite{Dix2023Measuring} & 2023 & ML model robustness testing, time series data quality & Propose a generic framework for systematically analyzing the impact of data quality issues on the performance of machine learning models by applying gradual perturbations to the original time series data. \\

   \cite{priestleySurveyDataQuality2023} & 2023 & DQ dimensions, DQ pipelines & Contribute a temporal mapping of important data quality requirements at different stages of the ML data pipelines within dimensions of intrinsic, contextual, representational, and accessibility. \\

   \cite{zha2023datacentric} & 2023 & Data-centric AI, AI benchmarks, data lifecycle & Present a comprehensive survey on the necessity of data-centric AI, detailing 3 primary data-centric goals (training data development, inference data development, and data maintenance), organizing existing literature from automation, collaboration, challenges, and benchmarks perspectives. \\

   \cite{barry2023impact} & 2023 & Training data quality, fairness metrics, supervised classification & Investigate how the quality of training data, specifically the presence of noise in labels and data, affects the fairness of supervised classification models across various algorithms and image classification datasets.\\

   \cite{ehrlingerSurveyDataQuality2022} & 2022 & DQ tools, DQ profiling, and monitoring & Investigate 13 data quality tools across data profiling, DQ measurement metrics, and continuous DQ monitoring, making up for practical implementation and highlighting a need for functional enhancements and critical discussion on widely accepted but under-implemented data quality metrics. \\

   \cite{B2020dataIoT} & 2020 & Big data and IoT, end-to-end data quality assessment, data-shared IoT platforms & Discuss the importance of data quality in the Internet of Things (IoT), emphasizing its direct impact on model results and business decisions. It highlights challenges in assessing data quality in IoT due to the growth and heterogeneity of IoT-connected devices, proposing the use of trust-based techniques combined with blockchain for secure end-to-end data quality assessment. \\

   \cite{Roh2021survey} & 2019 & Data collection, data labeling, deep learning & Examine data collection within ML from a data management perspective, which includes data acquisition (data discovery, data augmentation, and data generation) and data labeling (using existing labels, crowdsourcing, and weak labeling). It also surveys methods of improving labeling and models, such as transfer learning. \\

   \cite{Sidi2012survey} & 2012 & Data quality definition, dimensions, types, strategies, and techniques & Provide a comprehensive view of data quality definitions, DQ problems classification, 40 dimensions, and data quality strategies (data-driven and process-driven).\\

   \cite{batini2009methodologies} & 2009 & DQ dimensions, DQ assessment, and improvement & Provide a systematic and comparative description of methodologies for selecting, customizing, and applying DQ assessment and improvement techniques across various methodological phases, steps, strategies, techniques, data quality dimensions, types of data, and information systems.\\
     
\bottomrule
\end{tabularx}
\end{threeparttable}
\end{table*}

Table \ref{table:survey} presents existing surveys for evaluating and improving data quality. Surveys like \cite{Sidi2012survey} and \cite{batini2009methodologies} discussed DQ dimensions in a fundamental and general way, fitting common data analysis tasks. Others mainly paid attention to the metrics that affect ML models, such as completeness \cite{priestleySurveyDataQuality2023}, accuracy\cite{LiActiveLearning2024, priestleySurveyDataQuality2023}, robustness\cite{Dix2023Measuring}, fairness\cite{mohammed2024data}, \cite{barry2023impact}, and security\cite{B2020dataIoT}. In particular, there are papers focusing on certain DQ-improving techniques, such as active learning \cite{LiActiveLearning2024} and data generation \cite{lu2023mlSynthetic} or specific domains \cite{B2020dataIoT}. The existing studies on DQ for ML mainly discuss aspects as follows.

\begin{itemize}

    \item The impacts of DQ on ML models include the performance \cite{chenDataCuration2022, Chenphdthesis2022, ChenEvaluation2021}, bias \cite{yangFairerDatasetsFiltering2020, PAULLADA2021}, fairness \cite{KiDataCleaning2019}, and safety \cite{ChenSecuritySoftwareSystems2024}. 

    \item Scholars have put forward data quality frameworks, dimensions, and metrics based on different use cases \cite{Chenphdthesis2022, PLOTKIN2014127, widadQualityAnomalyDetection2023, heinrich2007measure, quaresmini2023data}. Open-source tools offer promising solutions for faster identification of DQ issues, facilitating current evaluation practices \cite{ehrlingerSurveyDataQuality2022}, \cite{ehrlingerAutomatedContinuousData2018}. They also lower barriers to DQ practices for a wider range of users, fostering collaboration and enabling further advancements. 

    \item Many studies have conducted experiments on real-world datasets to evaluate data quality. Corresponding techniques consist of LLMs for data generation \cite{li2024empowering, moller2023prompt, Yang2024llmgpt, ding2024data}, transfer learning \cite{ChenEvaluation2021}, and active learning \cite{LiActiveLearning2024}, etc. 

\end{itemize}


Nevertheless, most existing studies only include specific metrics without proposing corresponding dimensions. Our work fills this gap, and compiles a comprehensive set of DQ metrics applicable to the ML domain, providing more references for this field. Furthermore, research on DQ tools is relatively scarce. This paper presents the latest developments in tools from the past five years, identifies potential development directions, and explores how tools can leverage data-centric AI to play a more powerful role. The comparative analysis and roadmap for tool developers further facilitate the understanding of this field.

\section{Data Quality in Machine Learning}
\label{sec:2}
\subsection{Data quality dimensions}

Data quality is the comprehensive characterization and measurement of quantitative and qualitative properties of data. Based on application scenarios, the definition of data quality may involve different detailed descriptions. Wang and Strong defined data quality as ``data that are fit for use by data consumers" \cite{wangAccuracyWhatData1996}. Wand and Wang defined data quality as ``the quality of mapping between a real-world state and an information system state" \cite{wand1996anchoring}. In this research, data quality is defined as ``satisfying the needs and preferences of its users or tasks" or the capability of the data being ``fit for purpose" in ML \cite{ChenEvaluation2021}. A certain project can evaluate what combination suits the purposes best and design weighted metrics \cite{gongSurveyDatasetQuality2023, chenPracticalFrameworkEvaluating2019, foroniEstimatingExtentEffects2021, pradhanIdentifyingManagingData2023}. 


Wang and Strong put forward four dimensions: intrinsic, contextual, accessibility, and representational to form a data quality framework \cite{wangAccuracyWhatData1996}. Gong et al. summarized 20 papers from 2017 to 2023 and identified 8 quality dimensions, including completeness, self-consistency, timeliness, confidentiality, accuracy, standardization, unbiasedness, and ease of use \cite{gongSurveyDatasetQuality2023}. Cichy et al. pointed out the most common dimensions are completeness, accuracy, consistency, timeliness, and accessibility \cite{cichyOverviewDataQuality2019}. Data quality measurement metrics also include conformance, conformity, correctness, currency, duplicates (duplication), freshness, integrity, latency, plausibility, referential integrity, structure, uniformedness, uniqueness, validity, and other specific business rules, as summarized in Ehrlinger and Wöß’s paper \cite{ehrlingerSurveyDataQuality2022}. Informatica, the top ranking company in marketing share of data quality services, put forward 7 core data quality metric dimensions and their explanations \footnote{\url{https://www.informatica.com/resources/articles/what-is-data-quality.html}}. These dimensions can be further clustered to form a clearer understanding \cite{batiniDataInformationQuality2016a}. Schwabe et al. clustered the dimensions into measurement process, timeliness, representativeness, informativeness, and consistency \cite{schwabeMETRICframeworkAssessingData2024}. 

Dang et al. proposed a comprehensive taxonomy for data quality in NLP, including linguistic, semantic, anomaly, classifier performance, and diversity dimensions \cite{DangNLPDQ2024}. In the supervised fine-tuning of LLMs, the DQ dimensions of instruction data consist of instruction diversity, complexity, and prompt design \cite{wang2023dataforllm}. The primary goal of data-centric AI is not to improve the model training algorithm, but to improve the data pre-processing for better model accuracy \cite{whang2023data}. Based on this, more and more DQ evaluation frameworks and dimensions focus on the lifecycle of datasets \cite{gudivadaDataQualityConsiderations}, \cite{zha2023datacentric}, \cite{gongSurveyDatasetQuality2023}. The lifecycle of datasets includes data generation/initial stage, data acquisition/collection, data labeling, data cleaning, data processing and analysis, and DQ monitoring.

In Table \ref{table:1}, we summarize the frequently adopted data quality dimensions, definitions, examples, and metrics for reference. From the large quantity of DQ metrics, we have chosen the most pertinent ones for ML tasks. In this paper, we adopt the 4 dimensions put forward in \cite{wangAccuracyWhatData1996}:

\begin{itemize}
\item \textbf{Intrinsic} dimension can be assessed by measuring internal attributes or characteristics of data based on given references \cite{Stvilia2007}, \cite{elouataouiAdvancedBigData2022}. It also measures missing values and redundant cases.

\item \textbf{Contextual} dimension ensures that the data aligns with the needs and goals \cite{wangAccuracyWhatData1996} of the ML projects.

\item \textbf{Representational} dimension assesses the formats and structures of data, such as if the data is concisely and consistently represented, but also interpretable \cite{wangAccuracyWhatData1996}. 

\item \textbf{Accessibility} dimension evaluates the extent of obtaining either the entire or some portion of the data \cite{gudivadaDataQualityConsiderations}. Availability allows users to be able to use and share the data with safety controls.
\end{itemize}

\begin{table*}
\scriptsize
\caption{\centering DQ dimensions, metrics, descriptions, and examples}
\setlength{\tabcolsep}{3pt}
\renewcommand\arraystretch{1.1}

\begin{threeparttable}
\begin{tabularx}{\textwidth}{p{1.8cm}p{2.0cm}p{5.3cm}X}
\toprule
   \textbf{Dimension} & \textbf{Metrics}
    & \textbf{Description} & \textbf{Examples}\\ 
   \midrule

   & Correctness & A record in a dataset is free of errors \cite{Chenphdthesis2022}.  & Before starting a mailing campaign, the correctness of the attributes `‘postal code" shall be evaluated, and even small deviations shall be penalized because a deviation of only 1\% (the postal codes 80000 and 79200) hinders the delivery of a mailing \cite{heinrich2007measure}.\\
    & & Data is correctly labeled if it is a labeled record \cite{Chenphdthesis2022}. & In the medical domain, an informal phrase of `'lack of feeling" should be labeled as `‘numbness". \\

  Intrinsic &  Duplication & Measures if the same instances repeat in the dataset, especially in both the training and test datasets. & If a record in a medical concept training dataset is `‘Hunger – don’t want to eat", and there is exactly the same record in a test dataset, then the record is considered as an overlapped record in the two datasets. \\

   &  Trustworthiness & Defines how factual the source that provides the information is \cite{dong2015knowledgebased}. It can be subjectively evaluated, such as indicating the level on a scale \cite{GIL2007227}, \cite{feeneyImprovingCuratedWebData2014}, or the data can go through fact-check algorithms. & For a medical concept dataset, it should be obtained directly from the hospital's system, which undergoes regular data quality checks and is maintained according to industry standards. \\

   \hline

   & Class imbalance & Evaluates if the distribution of examples across the known classes is biased or skewed. & Most of the contemporary works on class imbalance fall into the imbalance ratios ranging from 1:4 up to 1:100. The imbalance ratio may range from 1:1000 up to 1:5000 for extreme class imbalance problems. \\

   & Completeness & A complete dataset should include as few missing values as possible. & A medical insurance dataset must include a customer's birthdate, otherwise the medical consumption forecast model performance will be hindered \cite{PLOTKIN2014127}.\\

  Contextual & Comprehensiveness & A dataset contains all representative samples from the population \cite{Chenphdthesis2022}.  &  In a medical text classification task, the training dataset should contain sufficient labeled medical texts covering all the conditions, symptoms, and treatments. \\

   & Unbiasedness & Refers to whether the data used for machine learning training has a distribution bias or historical bias \cite{gongSurveyDatasetQuality2023}. & Photo recognition software does not recognize the facial expressions of ethnic minorities, or electronic soap dispensers that do not respond to darker skin tones because the training image datasets have an insufficient representation of some geographic regions \cite{priestleySurveyDataQuality2023}. \\

   & Variety & Requires each validation dataset and the test dataset to contain a significant amount of new data compared to the corresponding training dataset. & The percentage of the overlapped data between a test/validation dataset and its corresponding training dataset should be as low as possible, such as less than 10\%. \\ 

   \hline

   & Conformity & Measures how much the data conforms to the conventions for capturing information in a certain manner, including machine-readable data structures and formats for capturing specific attributes \cite{Chenphdthesis2022}. 	& In a text classification task, a dataset of textual documents is labeled with sentiment (positive, negative, neutral). The labels should be encoded following a standardized set of categories, and all data processing, such as removing punctuation, converting text to lowercase, and tokenizing sentences, should be made to the whole dataset. \\

   Representational & Consistency & Requires data to be presented in the same format and to be compatible with previous data \cite{Chenphdthesis2022}. & In an image classification task, if one dataset uses pixel values in the range [0, 255], while another dataset scales pixel values to the range [0, 1], this will cause inconsistency in model training and predictions. \\

   \hline

   Accessibility 
   & Availability & High data availability ensures that data is readily accessible with defined user permissions for access and modifications. & In a healthcare ML application for diagnosing diseases from medical images, user entitlements are managed through strict access controls, allowing only authorized medical professionals and data scientists to access the images. \\

\bottomrule
\label{table:1}
\end{tabularx}
\end{threeparttable}
\end{table*}

\subsection{Impact of poor data quality on machine learning models}

Data quality impacts AI significantly, especially data-centric AI \cite{guo2020recommender} and LLMs. By examining label errors in the test sets of 10 of the most commonly used computer vision, natural language, and audio datasets, experiments have shown that pervasive label errors exist and destabilize ML benchmarks \cite{northcutt2021pervasive}. For large multi-modality models, ensuring high-quality, contextually relevant, and well-reasoned training data is crucial for leveraging the strengths of multiple models and producing superior results. 

Poor data quality means the datasets get low scores on one or even many DQ metrics as mentioned above. The quality of training data impacts the accuracy, reliability, and interpretation of ML models' results \cite{gahiMachineLearningDeep2021}, \cite{elouataouiAutomatedBigData2023}. For example, with many missing values in the previous purchasing records, the forecasting may deviate; With the absence of the data of a neighborhood, the ML model cannot learn useful information for this category, leading to biased and incorrect outcomes. Another example is that eye disease detection models trained on noise-free training data for high model performance can fail in detecting images with little variation. Sambasivan et al. concluded that this noise causes `‘data cascades", meaning using poor quality data in ML models and causing negative downstream effects \cite{NithyaDataWork2021}.

Specifically, deep learning techniques require large labeled datasets, underscoring the importance of sufficient and correctly labeled data \cite{Roh2021survey}, \cite{Wang2023labelquality}. If current labels are noisy, further labeling will not improve the model accuracy \cite{Sheng2008label}. Chen et al. evaluated label quality in publicly available datasets and found notable gender and age disparities in annotation quality \cite{ChenEvaluation2021}. They further enhanced AI training efficiency and performance by focusing on high-quality pseudo labels, resulting in a 33\%–88\% performance boost over entropy-based methods, with a cost of 31\% time and 4.5\% memory.

\subsection{Challenges in evaluating and improving data quality in ML}

Scholars come across challenges while developing and applying certain strategies or frameworks. Firstly, the lack of standardized metrics for DQ evaluation complicates the comparison of different approaches \cite{caiChallengesDataQuality2015}. Many DQ dimensions overlap, such as accuracy, correctness, and consistency. In a complex scenario, accuracy is affected by incomplete data, inconsistent data, etc., making the relationships of dimensions intertwined. Other overlapping dimensions, such as duplication and uniqueness \cite{JuddooDataGovernance2018}, time-related dimensions (timeliness, currency, and volatility), coverage, and comprehensiveness, also present the same concerns. Therefore, these should be determined after evaluating what combination fits the purpose most. 

Meanwhile, metrics need to be checked and updated according to developments in Ml research \cite{gongSurveyDatasetQuality2023}. Additional quality criteria must be considered based on specific ML tasks \cite{gongSurveyDatasetQuality2023}, or different phases of the ML development pipeline \cite{priestleySurveyDataQuality2023}. Applying existing methods to specific scenarios also requires modifications to metrics and designs. Codella et al. faced challenges in interpreting person-generated health and wellness data in one comprehensive and reasonable framework \cite{codellaDataQualityChallenges2018}. Li et al. also found using only two textual factors to evaluate review data quality was limiting and required an extended framework \cite{liHowTextualQuality2020}.

With large-volume and multi-source, multi-modal data,  it's important to continuously monitor and maintain DQ to ensure models remain effective \cite{ozonzeAutomatingElectronicHealth2023, ridzuanThematicReviewData2022}. Some DQ framework applications face time insufficiency by using traditional methods with a large workload \cite{wangAccuracyWhatData1996},\cite{zhangUnderstandingDetectingDefects2020}, \cite{abiyComparisonElectronicMedical2018}. The large scope and various types of data also add to the difficulties, making the evaluating algorithms and ML models more complex \cite{ridzuanThematicReviewData2022}, \cite{PansaraCultivatingDataQuality2023}. 

Additionally, further examination of the suggested DQ improvement plans is necessary to demonstrate their effectiveness in the results. However, the scope of the data quality framework sometimes includes only hypotheses of potential outcomes rather than further testing on multiple datasets. \cite{mutheeImpactRoutineData2018}. Sometimes the improvement plan can only work on a part of the datasets and serve as a small portion of data management \cite{vandenbergheDataQualityAssessment2017}. The time required and operational costs are also major constraints for choosing suitable plans \cite{kleindienstDataQualityImprovement2017}.

\section{Existing Open-Source Tools for Data Quality for Machine Learning}
\label{sec:3}

\subsection{Existing open-source tools for data quality}

Ehrlinger and Wöß established a three-fold selection strategy for existing open-source data quality tools: a systematic search, related surveys, and random online searches. They also set 6 exclusion criteria for the above results of 667 DQ tools to filter the final 13 targets (5 are open source). For example, for evaluation purposes, the tool has to be publicly available and offer a free trial \cite{ehrlingerSurveyDataQuality2022}. Based on this method, we summarize the existing mainstream DQ tools in Table \ref{table:2}, including applications and Python libraries. Table \ref{table:2} also presents the links, functions, and versions of these tools. The metrics column shows the main DQ metrics involved in data profiling, DQ issues detection, and transformation stages.

\begin{table*}
\caption{\centering A review of existing data quality tools}
\setlength{\tabcolsep}{3pt}
\renewcommand\arraystretch{1.3}

\begin{threeparttable}
\begin{tabularx}{\textwidth}{p{1.3cm}p{0.7cm}p{5.5cm}Xp{1.5cm}}
\toprule
   \textbf{Tools} & \textbf{Link} & \textbf{DQ metrics} & \textbf{Functions} & \textbf{Version}\\ 
   \midrule
   Kylo* & \href{https://kylo.io/}{Link} & Availability, Completeness, Consistency, Duplication & (1) Data profiling with automatic statistics (2) Data cleaning and standardization (3) Data monitoring & V0.10.1 on 2019-03-01 \\
   
   MobyDQ & \href{https://ubisoft.github.io/mobydq/}{Link} & Availability, Completeness, Correctness, Conformity & (1) DQ measurement and anomaly detection (2) Running automated data quality checks & Last update in 2020\\

   Apache Griffin* & \href{https://griffin.apache.org/\#about_page}{Link} & Completeness, Correctness, Conformity & (1) Defining DQ (2) DQ measuring with metrics with pre-defined DQ domain models & V0.6 on 2020-11-13 \\
   
   SQL Power Architect & \href{https://bestofbi.com/products/sql-power-architect-data-modeling/}{Link} & Consistency, Duplication, Conformity & (1) Data profiling (2) Database management (3) Visualization with mapping reports & V1.0.9 on 2020-12-23 \\

   Aggregate Profiler & \href{https://github.com/ru-fix/aggregating-profiler}{Link} & Consistency, Correctness, Trustworthiness, Conformity & (1) Data profiling (2) Data integration and cardinality check (3) Data generation and testing & V6.3.3 on 2021-01-20 \\

   YData Quality & \href{https://github.com/ydataai/ydata-quality}{Link} & Class imbalance, Comprehensiveness, Conformity, Consistency, Correctness, Duplication, Unbiasedness & (1) DQ evaluation throughout the multiple stages of a data pipeline development & V0.1.0 on 2021-09-23 \\

   DataCleaner & \href{https://datacleaner.github.io/}{Link} & Completeness, Correctness, Trustworthiness & (1) Data profiling and analytics (2) Data transformation (3) Data monitoring & V5.8.1 on 2022-02-09 \\ 

   WinPure & \href{https://winpure.com/}{Link} & Availability, Conformity, Consistency, Correctness, Duplication & (1) Data profiling and quality issues identification with AI (2) Data matching with domain knowledge (3) Data verification (4) Fast large-scale data transformation & V8.0 on 2023-06-29 \\

   SQL Power DQguru & \href{https://bestofbi.com/products/sql-power-dqguru-data-quality/}{Link} &  Correctness, Duplication, Trustworthiness, Conformity & (1) Data transformation (2) Data matching and merging (3) Data validation & V0.9.7 on 2023-08-17 \\

   Deequ* & \href{https://github.com/awslabs/deequ}{Link} &  Consistency, Correctness, Conformity & (1) Data profiling (2) Data testing (3) Automatic suggestion of constraints for DQ metrics & V2.0.6 on 2023-11-14 \\

   Dataedo & \href{https://dataedo.com/product/data-profiling}{Link} & Availability, Class imbalance, Comprehensiveness, Consistency, Duplication, Conformity & (1) Data sampling (2) Data profiling column distribution in terms of nullability and uniqueness, data types, and statistics & V24.1.2 on 2024-02-06 \\

   OpenRefine* & \href{https://openrefine.org/}{Link} & Availability, Conformity, Consistency, Correctness, Trustworthiness & (1) Data exploring and profiling (2) Data transformation and visualization (3) Data integration and reconciling: adding a platform, such as Wikidata, to fix spelling or variations in proper names, examining typos, whitespace, and others & V3.8 on 2024-02-21 \\

   Great Expectations & \href{https://github.com/great-expectations/great_expectations}{Link} & Consistency, Correctness, Trustworthiness, Conformity & (1) Data profiling (2) Data transformation and validation: ensuring the correctness of transformations or integration from another data source (3) Automating data quality checks over time: creating data docs for the validation results & V0.8.15 on 2024-05-29 \\

   Soda Core & \href{https://github.com/sodadata/soda-core}{Link} & Completeness, Correctness & (1) DQ checks (2) Data integration (3) Data monitoring with GPT & V3.3.5 on 2024-05-24 \\
   
   Ataccama ONE* & \href{https://www.ataccama.com/platform}{Link} & Availability, Completeness, Consistency, Duplication, Trustworthiness, Conformity, Variety &  (1) Data monitoring (2) Data profiling and cleansing (3) Detection with AI (4) Generating warnings about deviations with a DQ firewall & V15 on 2024-03-05 \\

   whylogs* & \href{https://github.com/whylabs/whylogs}{Link} & Class imbalance, Conformity, Consistency, Duplication & (1) Data profiling and visualization (2) Data merging and integration (3) Data and ML models monitoring & V1.4.0 on 2024-05-14 \\

   Evidently* & \href{https://www.evidentlyai.com/}{Link} & Availability, Class imbalance, Consistency, Comprehensiveness, Duplication, Unbiasedness, Variety & (1) Automatic DQ checks  (2) Interactive visualization (3) ML model performance, features, and DQ monitoring dashboard (4) Evaluation of data shifts and changes in the distribution and model predictions & V0.4.26 on 2024-06-07 \\
  
\bottomrule
\label{table:2}
\end{tabularx}

\begin{tablenotes}
      \small
      \item Note: The open-source version of Talend Studio was retired on January 31, 2024. Tools with a * indicate they support customized rules to evaluate data quality.
\end{tablenotes}

\end{threeparttable}
\end{table*}

\textbf{Talend Open Studio} was a free and open-source ETL (Extract, Transform, and Load) tool of Talend’s tool series \footnote{\url{https://www.talend.com/products/talend-open-studio/}}. DQ check is one of the segments of this tool. It enabled data cleaning, and data visualization with data drill-down capability, such as interactive charts. It supported normalizing spreadsheets or other datasets in different formats and automatically identified data types and potential errors.

\textbf{Kylo} provides a user interface to configure new data feeds including schema, security, validation, and cleansing, and the ability to wrangle and prepare visual data transformations using Spark as an engine. Its flexible data processing framework for building batch or streaming pipeline templates enables monitoring the whole data processing procedure. 

\textbf{MobyDQ} helps data engineers automate DQ checks on data pipelines. Its DQ framework includes 5 aspects: anomaly detection, completeness, freshness, latency, and validity. It can work on various data sources, such as MySQL, PostgreSQL, Teradata, Hive, Snowflake, and MariaDB. 

\textbf{Apache Griffin} is a DQ solution for big data, which supports both batch and streaming modes. It offers domain models that cover most general DQ problems. It also helps users define their quality criteria and enables users to implement their specific functions. 

\textbf{SQL Power Architect} is a data modeling and profiling tool. It can provide a complete view of all required database structures and expedite every aspect of the data warehouse design. The auto-layout tree view of the schemas generates information about the data size, maximum and minimum values, frequency, etc. It stores the origin of each column and can automatically generate the source-to-target data mappings. 

\textbf{Aggregate Profiler} is a data profiling and data preparation tool. It offers advanced data profiling methods, such as metadata discovery, anomaly detection, and pattern matching. In addition, it supports many tasks beyond data profiling, including masking, encryption, governance, integration, reporting, and dummy data creation for testing.

\textbf{YData Quality} is an open-source Python library for assessing DQ issues throughout the multiple stages of data pipeline development. It mainly evaluates bias and fairness, data expectations, data relations, data drifts, duplicates, labeling, missing data, and erroneous data. It also supports low-code commands.

\textbf{DataCleaner} is a data profiling tool for discovering and analyzing data quality with monitoring. It allows customized cleansing rules and composing them into different use scenarios or target databases. It supports simple search rules, regular expressions, pattern matching, and other custom transformations. Data composition and conversion are available. 

\textbf{WinPure} is a data matching and cleansing tool. Software training, tutorials, and guideline are provided to enhance user experience. It can clean, correct, standardize, and transform data. All settings can be saved and used on other similar datasets. Its data profiling and quality issues identification provide over 30 different statistics highlighting potential DQ issues. Data matching is equipped with domain knowledge. It can also check the validity and deliverability of any global mailing address, automatically correcting and adding all missing address elements. 

\textbf{SQL Power DQguru} helps cleanse data, validate and correct addresses, identify and remove duplicates, and build cross-references between source and target tables. It displays color-coded match diagrams on a comprehensive match validation screen, and data conversion workflow through the visualization of an intuitive transformation process.

\textbf{Deequ} defines `‘unit tests for data", and measures data quality in large datasets. It supports querying computed metrics from a metrics repository. It can detect anomalies over time and automatically suggest specific rules, and incremental metrics computation on growing data. It can work on tabular data like CSV files, database tables, logs, flattened JSON files, and anything that fits a Spark data frame. 

\textbf{Dataedo} can extract data samples, which help users learn about the dataset. The data profiling function enables the calculation of common measures like row numbers and the percentage of distinct and empty rows. It also shows statistics with visualizations. It supports data documentation and sharing with teams.

\textbf{OpenRefine} is a data cleaning, transforming, and extending tool. It provides data profiling and an overview of data types, with precise conversion and formatting, using expressions and arguments identifying key columns. It enables data transformation through common and customized methods, including clustering, pulling data from the web, reconciling, and writing expressions. MetricDoc is its extension with an interactive environment. It assesses DQ with customizable, reusable quality metrics and provides immediate visual feedback, to facilitate interactive navigation and determine the causes of quality issues. 

\textbf{Great Expectations} helps with DQ testing, documentation, and profiling. Some key features are seamless operation, fast results with large volume data, a flexible, extensible, and human-readable vocabulary, data contracts support, and easy collaboration.

\textbf{Soda Core} is a Python library that enables finding insufficient data. It supports data testing in development, pipelines, and definitions in human-readable language. It can integrate frameworks to most databases through extensive Python and REST APIs, and the reports can be shared with others by email and Teams to get quality issues alerts.

\textbf{Ataccama ONE} is a data profiling and analysis tool. The free version is no longer available and is provided only as part of the Ataccama ONE platform. Ataccama ONE offers data catalog, reference data management, data integration, and data story functions. This tool is also AI-driven. It works on automating tasks and developing models, providing real-time issue identification with all DQ metrics in an integrated data catalog. It has been applied in diverse industries including healthcare, transportation, banking, retail, telecom, and government.  

\textbf{Whylogs} is a data logging library for machine learning models and data pipelines. It captures key statistical properties of data, such as the distribution, the number of missing values, and a wide range of configurable custom metrics. It can track data distributions, DQ issues for ML experiments, and model performance over time.

\textbf{Evidently} is an open-source Python library for data scientists and ML engineers to evaluate, test, and monitor ML models and data quality. It works with tabular, text data, and embeddings in NLP and LLM tasks. It supports building a custom report from individual metrics. The highlight is its monitoring ability throughout the ML lifecycle by tracking model features over time. Powered by AI, it can run data profiling with a single line of code, and solve nulls, duplicates, and violations in production pipelines.

\subsection{Comparative analysis of tools based on features, usability, effectiveness, and others}

Our comparative analysis of the open-source tools contains their difference in functions, the DQ metrics, as well as their usability and interface. Previous work has provided a holistic view of tool evaluation. Ehrlinger and Wöß evaluated 13 data quality tools from 3 functional dimensions; data profiling, data quality measurement (metrics), and automated data quality monitoring \cite{ehrlingerSurveyDataQuality2022}. Based on the classification put forward by Abedjan et al. \cite{abedjanDataProfiling2019}, Ehrlinger and Wöß put forward 5 data profiling sub-categories (namely: cardinalities; value distributions; data types, patterns, and domains; dependencies; and advanced multi-columns profiling), as well as according test results, and the measurement metrics included accuracy, completeness, consistency, and timeliness. As for the function of data monitoring, it is more difficult to define and does not have a common understanding. They included this dimension with some new evidence discovered in existing tools. At the same time, they also investigated vendors, data cleansing, and technical features as supplements. From these perspectives, we formed our comparative analysis in the following figures and summaries. 

\begin{figure}
    \centering
    \includegraphics[width=1\linewidth]{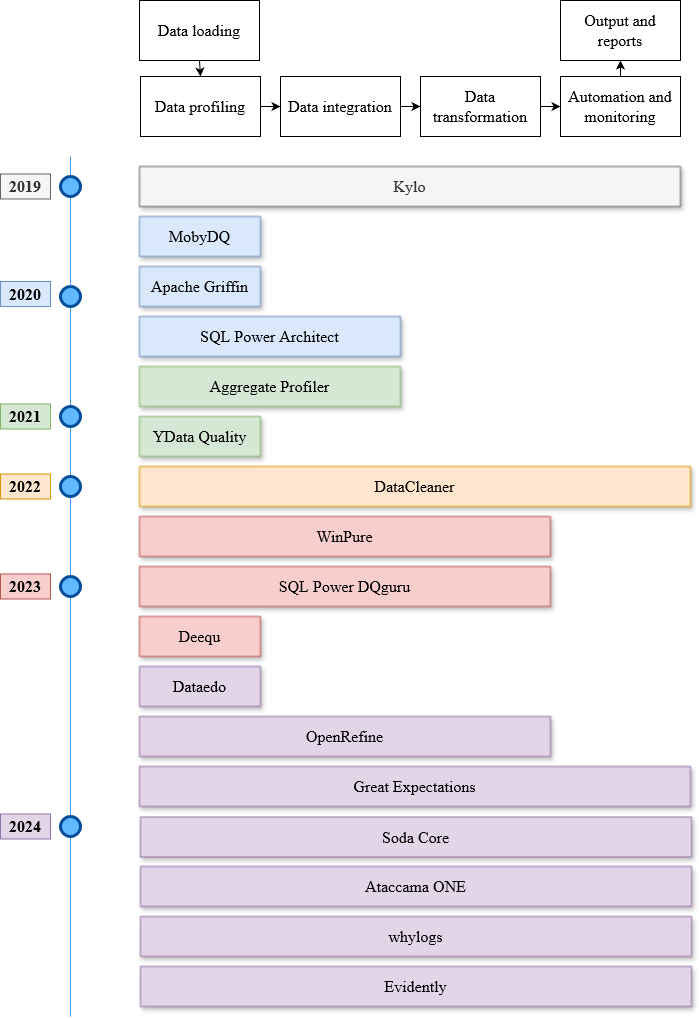}
    \caption{Evolution of DQ evaluation/improvement tools across functions over time. The 6 core functions are data loading, data profiling, data integration, data transformation, automation and monitoring, and output and reports. Every tool supports the loading and output functions so the middle four remain for discussion. The length of each tool shows its coverage of the functions and the color indicates the last year that the tool was updated.}
    \label{fig:evolution}
\end{figure}

(1) The overall trend. Figure \ref{fig:evolution} visualizes the latest update time of each tool and the functions to which they upgrade. Most tools updated in 2024 have evolved to the functions of automation and monitoring, indicating the new trending direction of developing data quality tools.

(2) Data profiling function. Data profiling aims to describe a dataset and provide insights. It is the first step for the users to have an overview of the data, get to know data quality issues, and decide on corresponding fixing strategies. As shown in Figure \ref{fig:evolution}, most tools have this core function and put it at the forefront. DataCleaner, Ataccama ONE, and WinPure enable more features, such as frequency statistics, duplication analysis, data pattern discovery, drill-through analysis, etc. Some tools focus on data profiling and DQ issues detection, such as MobyDQ, YData Quality, Deequ, and Dataedo. 

(3) Data integration function. This helps maintain consistency when the customer wants to merge data from a different source. Many tools support this demand. Moreover, Ataccama ONE supports large-load data integration with seamless performance and continuous data quality checks during the process.

(4) Data transforming function. Transforming data means taking actions to fix the issues presented in the data profiling or discovery stage, and may include data cleaning, matching, merging, and de-duplicating tasks. WinPure, OpenRefine, and SQL PowerDQguru enable a comprehensive series of transforming abilities.

(5) Automation and monitoring. Once the rules for profiling, transforming, integration, and other tasks are set, some tools can automatically re-activate the workflow when new data is coming and generate up-to-date reports. Winpure, Ataccama ONE, Soda Core, and Evidently are the AI-embedded tools that can facilitate this and improve cost and time efficiency. 

\begin{figure*}[ht]
    \centering
    \includegraphics[width=1\linewidth]{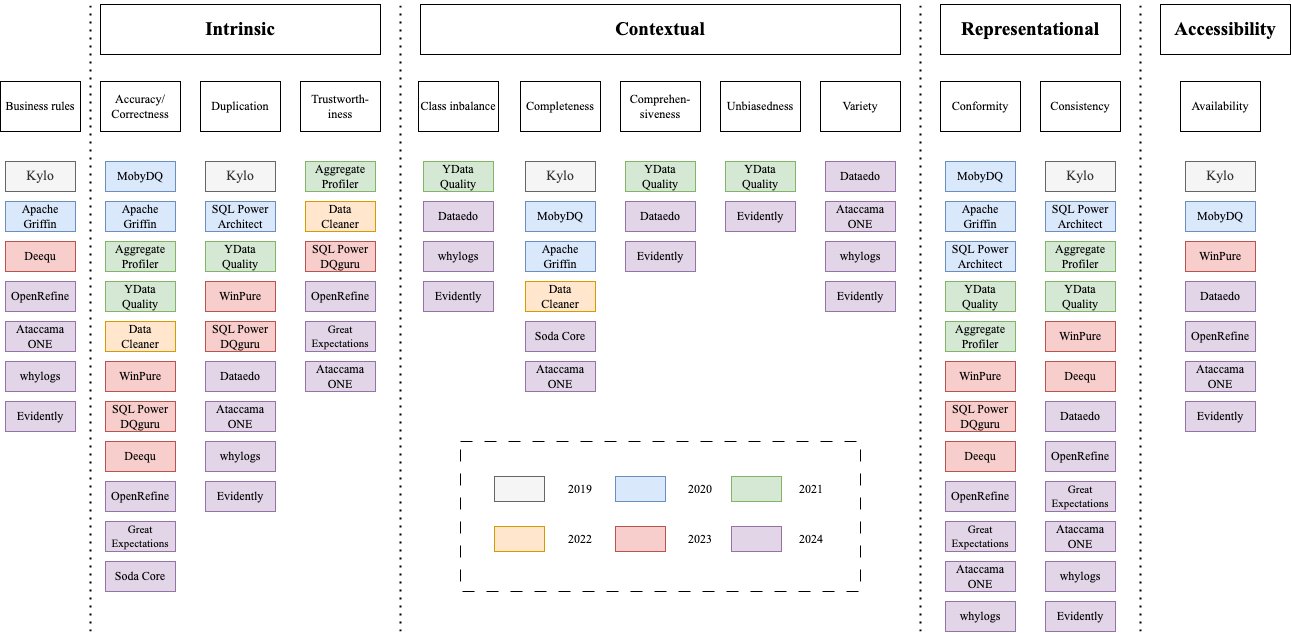}
    \caption{DQ dimensions, metrics, and corresponding tools. It showcases 4 dimensions and 12 DQ metrics in the first and second rows. Beneath each one, corresponding tools are listed, indicating their evaluation focus on the specific metrics and dimensions. The color of each tool represents the last year that the tool was updated as shown in the middle bottom corner. Business rule is listed at the left as an additional aspect as many tools support customized rules.}
    \label{fig:DimensionandTool}
\end{figure*}

(6) The most adopted data quality metrics. DQ tools usually do not present more than 5 metrics, which might be explained by reducing technical complexity. YData Quality, Ataccama ONE, whylogs, and Evidently support more DQ metrics checks than others. From Figure \ref{fig:DimensionandTool}, the most frequent metrics are consistency, conformity, accuracy/correctness, and duplication. The least are class imbalance, comprehensiveness, unbiasedness, and variety, which indicates current tools focus more on general DQ evaluation tasks and relatively less on specific ML ones. Among all the tools, YData Quality, Ataccama ONE, whylogs, and Evidently are designed to be more tailored to ML tasks while others focus more on general big data DQ tasks.

(7) The user interface. From the release date and the latest version of each tool, we can see that some tools have not been updated for a long time, like MobyDQ, Apache Griffin, and SQL Power Architect. Their user interfaces are relatively simple, lacking in design, interaction, or enough information a user may need. On the other hand, Ataccama ONE and Evidently demonstrate the best user experience, with useful guides, clear descriptions of features, example cases, and an easy-to-navigate, well-designed website. Moreover, YData Quality and Evidently support low-code commands and make the tool more user-friendly.

\subsection{Strengths and limitations of current tools}

Current data quality tools have already established a framework for evaluation and improvement. Most tools highlight their core functionalities, such as data profiling, transformation, integration, or testing, and integrate various functions to streamline the process, offering customers choices to choose what fits their purposes most. From the comparative analysis, we can see 4 tools – Winpure, Ataccama ONE, Soda Core, and Evidently – have already stepped out to integrate AI and GPT technology into the modeling, rules suggestion, and monitoring tasks. This development also makes data quality checks more friendly to non-tech users, such as product managers and business owners. With heated discussions about these cutting-edge transformations, this could be a direction for the tools to update and empower their strengths.

In the meantime, there are still some limitations. Firstly, the complexity of technical terms and downloading may discourage users. For example, many tools only support setting up by codes and have specific environment restrictions. Further, not many tools enable customized data quality check rules or revising the current rules to fit a certain request. This could be a setback in a real-world scenario. 

Current tools have not made the data quality metrics clear. Some tools combine several dimensions and metrics and do not use the standardized definition. Further, some adopted metrics are used generally for most data analysis, and only a few tools support evaluating DQ issues specific to ML tasks. Designing DQ tools in ML is a future direction, which should be able to include DQ metrics in ML and monitor how DQ improvement help the model performance.

Regarding the growing volume of data processing, the developers of DQ tools should also think about how to handle larger datasets while monitoring data continuously. Julio et al. proposed a data quality model called DQ-MAN to address the high data quality demands in air quality monitoring systems \cite{buelvasDQMANToolMultidimensional2023}, but this tool was limited by the predetermined time window. Qualle \cite{ehrlingerAutomatedContinuousData2018}, proposed by Lisa et al. as an upgrade, can perform tasks continuously over time. However, the challenges appear as the tool is applied in practical cases with different levels of data volatility, such as the conformance of Wikipedia data, daily sales data, and continuous measurement of stock data. They found that the time progression of certain data quality measures might deviate from intuitive understanding, so more explanations and improvement methods need to be explored in the future.

\section{Roadmap for Developing Data Quality Tools for Machine Learning}
\label{sec:4}

\subsection{Steps of creating tools for data quality evaluation and improvement}

Developers need to form a comprehensive framework and set clear objectives and scopes to create DQ evaluation tools that cater to customer needs and fill the gap of current limitations. 

(1) Understand the background. The background includes the knowledge of general data quality definitions, dimensions, metrics, existing limitations, and the market expectations of DQ tools in a particular domain, such as NLP tasks and their targeting goals, etc.

(2) Define the scope and key features. It is crucial to select the range of data quality metrics that the tool can check and improve upon. Some questions can be: Will it involve the mainstream metrics, like other tools? What are other particular metrics that matter to this development? Does it pay attention to ML tasks, textual data in LLM, or other general dimensions? What kind of data does it process, like structured or unstructured? What type of sources and how large data volume does it support? What are the key functions of this tool?

(3) Implement technology stack, functions, connectors, and metrics. Based on the requirements, choose the suitable languages, libraries, and frameworks to build the architecture, such as Java, Python, Apache Spark, MySQL, and cloud platforms. Connection to APIs should also be considered. While defining metrics, the general metrics should be offered as suggestions and there should also be room for users to make changes.

(4) Create user interfaces and design websites. For the tools facing customers including non-tech users, the user interfaces should be friendly and simple, with a clear introduction to the functions, hands-on guides, video tutorials, and industry examples. It should also be intuitive with the website presenting dashboards and visualizations \cite{RangineniAnalysis2023}, and highlighting the difference from other tools. For example, Mang et al. designed a user-centered graphical user interface (GUI) for a data quality evaluation tool. The interface is interactive, low-code, and provides detailed investigations of possible DQ irregularities \cite{mangDQAguiGraphicalUser2022}. Priestley et al. also considered dashboards and visual aids important for data inspection and sanity checks, which helps data exploration with all kinds of figures including pair plots, distributions, correlations, histograms, or heatmaps \cite{priestleySurveyDataQuality2023}.

(5) Establish documentation, community support, stories, and other engagement. After releasing it as an open-source tool, it is crucial to answer users' questions, encourage support, and foster a beneficial communication environment. Additionally, extracting insightful feedback and evaluating the system's validity, reliability, and generalizability are important. 

\subsection{Key components and functionalities in a data quality tool}

\begin{figure*}[ht]
    \centering
    \includegraphics[width=0.75\linewidth]{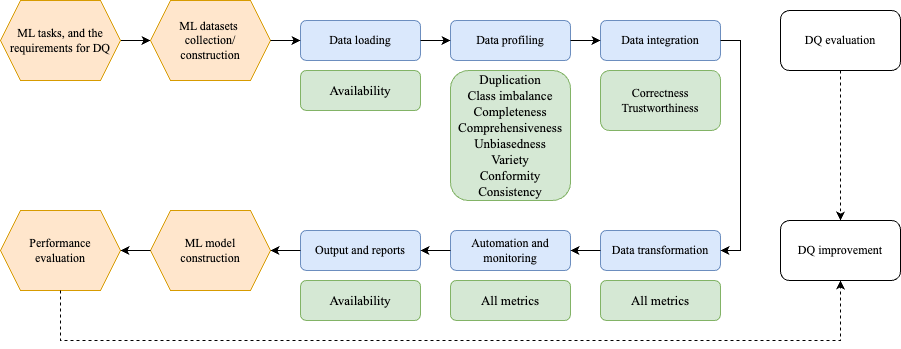}
    \caption{Workflow of the data quality evaluation and improvement. The orange figures represent the parts of ML model constructions, the blue ones are the functions of DQ tools, and the metrics are shown in green boxes below each step.Specific ML tasks set certain DQ requirements, leading to dataset collection and subsequent evaluation and improvement. Finally, model performance reflects the effectiveness of the DQ improvement process.}
    \label{fig:workflow}
\end{figure*}

The functions are the core of a data quality tool. In this section, we illustrate DQ evaluation and improvement steps by Figure \ref{fig:workflow}. The descriptions provide the developers with directions for designing DQ tools.  

(1) Data loading. The tool may support various data sources, such as databases, flat files, APIs, data lakes, and streaming platforms, to ensure compatibility and usability. Other than that, data volume is another concern. The tool may suggest data sampling to speed up the evaluation process and provide quick insights.

(2) Data profiling. The basic function should include summary statistics and visualization for columns or customized indicators. Users also can define and incorporate quality metrics that align with their data quality objectives. These metrics could include common metrics, such as correctness, completeness, duplication, uniqueness, and consistency. The overall data quality can be measured by weighted metrics. Incorporating ML models for more advanced data quality evaluation can be useful, such as classification models for anomaly detection or regression models for data imputation. 

(3) Data integration. Integrating data requires the evaluation of other sources of data, including various formats, and correctness. Reference datasets or knowledge in a specific domain may be needed. This could include appending demographic data, geocoding addresses, consulting reports, relevant datasets or databases, and third-party APIs, such as Wikidata. In ML, alignment and co-learning are the two relevant integration techniques that can be adopted in data integration \cite{whangDataCollectionQuality2023a}. Hanlon et al. provided a detailed design. The user can create the dataset interlinks at the schema level by selecting relationship-type and link-type, and adding a narrative provenance, with an instruction of a visual interpretation of the integrated data on the tool \cite{hanlonEffectiveUserInterface2021}.

(4) Data transformation. Users can determine what evaluation metrics to apply, or customize feasible business rules, enter qualitative dimensions and scores, and then determine the corresponding conversions to improve the data quality in each dimension. As Batini et al. pointed out, contextual knowledge, quality objectives, and budget constraints are the main concerns during the selection of strategies and techniques \cite{batini2009methodologies}. To ensure transparency and auditability, there is also a need to maintain a record of data lineage and track the transformations applied to the data from its source to its final form. By comparing the original and improved data, users can confirm the correctness of the functionality and check the effectiveness in evaluating data quality. Users can also choose the work modes, such as batch or streamline. 

(5) Automation and monitoring. Data monitoring is to automatically implement the defined metrics for the updated dataset and generate new reports, facilitating improvement plans according to the evaluation results. When users confirm the effectiveness and accuracy of the DQ evaluation and transformation results, the whole workflow can be saved and automated when there is new data. Users can also set the frequency of monitoring reports. This process can re-evaluate the metrics in previous funcations.

In ML, frequent changes in input and unexpected outliers can crush models, increasing the need for re-training. Avanade, an IT consulting services provider in collecting and analyzing data from Office 365 platforms, faced the impact of fluctuating input data on ML models. It implemented Great Expectations to profile, validate, and monitor the data quality. The outliers will be manually checked and revised, and then the transformation results undergo continuous validation and monitoring to ensure data quality at every processing stage. The transparent data logs also allow tracking of the metrics and results over time \footnote{How Avanade uses GX to detect data drift from upstream model changes in machine learning pipelines: https://greatexpectations.io/case-studies/how-avanade-uses-gx-to-detect-data-drift-from-upstream-model-changes-in}.

ML can increase the granularity and speed of DQ monitoring workflow and assist human laborers in detecting anomalies with fewer errors \cite{Wachirapusitan_2023}, \cite{DolatshahCrowdsourced2018}. Thresholds and alerts can be set based on historical records and domain statistics, and adjusted by dynamic interventions. One of the successful stories is that Ataccama ONE implemented 33 rules to monitor patients' data since hospitals need real-time decision support \footnote{US Healthcare Provider: https://www.ataccama.com/success-story/us-healthcare-provider}. The clinical workflow can generate alerts to any potential issue of data quality since it may influence the patient's outcomes. Similarly, Deequ can automatically analyze the statistics of the training data, check if it violates the defined data schema, provide a warning, and recommend improvement strategies \cite{LWAKATARE2020}.

(6) Output and reports. The final output can be a dashboard consisting of DQ metrics, and summaries of each function mentioned above, with highlighted DQ issue warnings and possible solutions. The tracking log can be showcased in a time-series manner, indicating the effectiveness of the improvement intuitively and facilitating the storytelling. The reports should support various formats and are only accessible to uses with control.

\section{Conclusion and Future Directions}
\label{sec:5}

This paper provides a comprehensive overview of DQ dimensions, examples, and metrics, and identifies challenges in evaluating DQ in ML. These challenges include profiling excessive data and monitoring frequently updated data. The functionalities of the latest open-source tools available on the market can fill certain gaps. Additionally, this paper presents a workflow for designing open-source tools by outlining the process of data profiling, integration, and transformation. Emphasizing the integration of AI trends in DQ management, this study fosters the creation of more effective tools and optimizes existing ones to better meet DQ standards.



AI-powered solutions are playing important roles in enhancing DQ evaluation \cite{muhammadImprovingLargeData2023}, \cite{dhoopatiEnhancingEnterpriseApplication2023}. Specifically, data augmentation methods utilizing GPT and LLMs can facilitate DQ improvement \cite{MAHARANADA202291}. This is achieved by increasing both the amount and variation of data \cite{ChenEmpiricalDA2023}, \cite{LIDataAugmentation202271}. Generative data augmentation can increase the volume of synthetic training examples, which can be informative and of high quality \cite{YangGDA2020}. Dai et al. proposed AugGPT to augment text data, addressing the issue of limited training samples by generating multiple variations of each sentence with similar concepts but different semantics. This approach can improve data consistency and the robustness of few-shot classification tasks \cite{dai2023auggpt}.

Furthermore, as AI powers more technological ideas with low-code requirements \cite{SUNDBERG2023}, user-friendly, self-service DQ tools are expected to become more prevalent, enabling non-technical users to efficiently conduct DQ evaluation and improvement tasks. These platforms may feature intuitive interfaces, integrated dashboards, guided workflows, and automated recommendations. For example, Cheng et al.'s interactive framework facilitates early error detection and improves data-centric workflows, making it accessible even for developers without strong ML backgrounds \cite{ChengToolSupport2023}.

\section*{Acknowledgments}

The authors would like to thank Marie Bloechle at the University of North Texas for editing the language and writing of the paper. 

\bibliographystyle{IEEEtran}
\bibliography{reference}

\begin{thebibliography}{10}
\providecommand{\url}[1]{#1}
\csname url@samestyle\endcsname
\providecommand{\newblock}{\relax}
\providecommand{\bibinfo}[2]{#2}
\providecommand{\BIBentrySTDinterwordspacing}{\spaceskip=0pt\relax}
\providecommand{\BIBentryALTinterwordstretchfactor}{4}
\providecommand{\BIBentryALTinterwordspacing}{\spaceskip=\fontdimen2\font plus
\BIBentryALTinterwordstretchfactor\fontdimen3\font minus \fontdimen4\font\relax}
\providecommand{\BIBforeignlanguage}[2]{{%
\expandafter\ifx\csname l@#1\endcsname\relax
\typeout{** WARNING: IEEEtran.bst: No hyphenation pattern has been}%
\typeout{** loaded for the language `#1'. Using the pattern for}%
\typeout{** the default language instead.}%
\else
\language=\csname l@#1\endcsname
\fi
#2}}
\providecommand{\BIBdecl}{\relax}
\BIBdecl

\bibitem{chenDataCuration2022}
N.~Tran, H.~Chen, J.~Bhuyan, and J.~Ding, ``Data curation and quality evaluation for machine learning-based cyber intrusion detection,'' \emph{IEEE Access}, vol.~10, pp. 121\,900--121\,923, 2022.

\bibitem{FENZA2021107366}
G.~Fenza, M.~Gallo, V.~Loia, F.~Orciuoli, and E.~{Herrera-Viedma}, ``Data set quality in machine learning: {{Consistency}} measure based on group decision making,'' \emph{Applied Soft Computing}, p. 107366, 2021.

\bibitem{RangineniAnalysis2023}
S.~Rangineni, ``An analysis of data quality requirements for machine learning development pipelines frameworks,'' \emph{International Journal of Computer Trends and Technology}, vol.~71, pp. 16--27, 08 2023.

\bibitem{GeigerGarbageInGarbageOut2020}
\BIBentryALTinterwordspacing
R.~S. Geiger, K.~Yu, Y.~Yang, M.~Dai, J.~Qiu, R.~Tang, and J.~Huang, ``Garbage in, garbage out? do machine learning application papers in social computing report where human-labeled training data comes from?'' in \emph{Proceedings of the 2020 Conference on Fairness, Accountability, and Transparency}, ser. FAT* '20.\hskip 1em plus 0.5em minus 0.4em\relax New York, NY, USA: Association for Computing Machinery, 2020, p. 325–336. [Online]. Available: \url{https://doi.org/10.1145/3351095.3372862}
\BIBentrySTDinterwordspacing

\bibitem{rao2020quality}
R.~Rao, S.~Rao, E.~Nouri, D.~Dey, A.~Celikyilmaz, and B.~Dolan, ``Quality and relevance metrics for selection of multimodal pretraining data,'' in \emph{Proceedings of the IEEE/CVF Conference on Computer Vision and Pattern Recognition Workshops}, 2020, pp. 956--957.

\bibitem{ChenEvaluation2021}
H.~Chen, J.~Chen, and J.~Ding, ``Data evaluation and enhancement for quality improvement of machine learning,'' \emph{IEEE Transactions on Reliability}, vol.~70, no.~2, pp. 831--847, 2021.

\bibitem{budachEffectsDataQuality2022a}
L.~Budach, M.~Feuerpfeil, N.~Ihde, A.~Nathansen, N.~Noack, H.~Patzlaff, F.~Naumann, and H.~Harmouch, ``The {{Effects}} of {{Data Quality}} on {{Machine Learning Performance}},'' Nov. 2022.

\bibitem{joshi2024dataefficient}
S.~Joshi, A.~Jain, A.~Payani, and B.~Mirzasoleiman, ``Data-efficient contrastive language-image pretraining: Prioritizing data quality over quantity,'' 2024.

\bibitem{gudivadaDataQualityConsiderations}
V.~N. Gudivada, A.~Apon, and J.~Ding, ``Data {{Quality Considerations}} for {{Big Data}} and {{Machine Learning}}: {{Going Beyond Data Cleaning}} and {{Transformations}},'' \emph{International Journal on Advances in Software}, vol.~10, pp. 1--20, 07 2017.

\bibitem{GuptaDataQuality2021}
\BIBentryALTinterwordspacing
N.~Gupta, S.~Mujumdar, H.~Patel, S.~Masuda, N.~Panwar, S.~Bandyopadhyay, S.~Mehta, S.~Guttula, S.~Afzal, R.~Sharma~Mittal, and V.~Munigala, ``Data quality for machine learning tasks,'' in \emph{Proceedings of the 27th ACM SIGKDD Conference on Knowledge Discovery \& Data Mining}, ser. KDD '21.\hskip 1em plus 0.5em minus 0.4em\relax New York, NY, USA: Association for Computing Machinery, 2021, p. 4040–4041. [Online]. Available: \url{https://doi.org/10.1145/3447548.3470817}
\BIBentrySTDinterwordspacing

\bibitem{zha2023datacentric}
D.~Zha, Z.~P. Bhat, K.-H. Lai, F.~Yang, Z.~Jiang, S.~Zhong, and X.~Hu, ``Data-centric artificial intelligence: A survey,'' 2023.

\bibitem{ehrlingerSurveyDataQuality2022}
L.~Ehrlinger and W.~W{\"o}{\ss}, ``A {{Survey}} of {{Data Quality Measurement}} and {{Monitoring Tools}},'' \emph{Frontiers in Big Data}, p. 850611, Mar. 2022.

\bibitem{mohammed2024data}
S.~Mohammed, H.~Harmouch, F.~Naumann, and D.~Srivastava, ``Data quality assessment: Challenges and opportunities,'' \emph{arXiv preprint arXiv:2403.00526}, 2024.

\bibitem{LiActiveLearning2024}
\BIBentryALTinterwordspacing
N.~Li, Y.~Qi, C.~Li, and Z.~Zhao, ``Active learning for data quality control: A survey,'' \emph{J. Data and Information Quality}, may 2024, just Accepted. [Online]. Available: \url{https://doi.org/10.1145/3663369}
\BIBentrySTDinterwordspacing

\bibitem{lu2023mlSynthetic}
Y.~Lu, M.~Shen, H.~Wang, X.~Wang, C.~van Rechem, and W.~Wei, ``Machine learning for synthetic data generation: a review,'' \emph{arXiv preprint arXiv:2302.04062}, 2023.

\bibitem{Dix2023Measuring}
M.~Dix, G.~Manca, K.~C. Okafor, R.~Borrison, K.~Kirchheim, D.~Sharma, K.~Chandrika, D.~Maduskar, and F.~Ortmeier, ``Measuring the robustness of ml models against data quality issues in industrial time series data,'' in \emph{2023 IEEE 21st International Conference on Industrial Informatics (INDIN)}, 2023, pp. 1--8.

\bibitem{priestleySurveyDataQuality2023}
M.~Priestley, F.~O'donnell, and E.~Simperl, ``A {{Survey}} of {{Data Quality Requirements That Matter}} in {{ML Development Pipelines}},'' \emph{Journal of Data and Information Quality}, no.~2, pp. 1--39, Jun. 2023.

\bibitem{barry2023impact}
A.~Barry, L.~Han, and G.~Demartini, ``On the impact of data quality on image classification fairness,'' in \emph{Proceedings of the 46th International ACM SIGIR Conference on Research and Development in Information Retrieval}, 2023, pp. 2225--2229.

\bibitem{B2020dataIoT}
J.~Byabazaire, G.~O'Hare, and D.~Delaney, ``Data quality and trust: A perception from shared data in iot,'' in \emph{2020 IEEE International Conference on Communications Workshops (ICC Workshops)}, 2020, pp. 1--6.

\bibitem{Roh2021survey}
Y.~Roh, G.~Heo, and S.~E. Whang, ``A survey on data collection for machine learning: A big data - ai integration perspective,'' \emph{IEEE Transactions on Knowledge and Data Engineering}, vol.~33, no.~4, pp. 1328--1347, 2021.

\bibitem{Sidi2012survey}
F.~Sidi, P.~H. Shariat~Panahy, L.~S. Affendey, M.~A. Jabar, H.~Ibrahim, and A.~Mustapha, ``Data quality: A survey of data quality dimensions,'' in \emph{2012 International Conference on Information Retrieval \& Knowledge Management}, 2012, pp. 300--304.

\bibitem{batini2009methodologies}
C.~Batini, C.~Cappiello, C.~Francalanci, and A.~Maurino, ``Methodologies for data quality assessment and improvement,'' \emph{ACM computing surveys (CSUR)}, vol.~41, no.~3, pp. 1--52, 2009.

\bibitem{Chenphdthesis2022}
H.~Chen, ``Data quality evaluation and improvement for machine learning,'' Ph.D. dissertation, University of North Texas, USA, 05 2022, aAI30182981.

\bibitem{yangFairerDatasetsFiltering2020}
K.~Yang, K.~Qinami, L.~{Fei-Fei}, J.~Deng, and O.~Russakovsky, ``Towards {{Fairer Datasets}}: {{Filtering}} and {{Balancing}} the {{Distribution}} of the {{People Subtree}} in the {{ImageNet Hierarchy}},'' in \emph{Proceedings of the 2020 {{Conference}} on {{Fairness}}, {{Accountability}}, and {{Transparency}}}, Jan. 2020, pp. 547--558.

\bibitem{PAULLADA2021}
A.~Paullada, I.~D. Raji, E.~M. Bender, E.~Denton, and A.~Hanna, ``Data and its (dis)contents: {{A}} survey of dataset development and use in machine learning research,'' \emph{Patterns}, no.~11, p. 100336, 2021.

\bibitem{KiDataCleaning2019}
\BIBentryALTinterwordspacing
K.~H. Tae, Y.~Roh, Y.~H. Oh, H.~Kim, and S.~E. Whang, ``Data cleaning for accurate, fair, and robust models: {A} big data - {AI} integration approach,'' \emph{CoRR}, vol. abs/1904.10761, 2019. [Online]. Available: \url{http://arxiv.org/abs/1904.10761}
\BIBentrySTDinterwordspacing

\bibitem{ChenSecuritySoftwareSystems2024}
\BIBentryALTinterwordspacing
H.~Chen and M.~A. Babar, ``Security for machine learning-based software systems: A survey of threats, practices, and challenges,'' \emph{ACM Comput. Surv.}, vol.~56, no.~6, feb 2024. [Online]. Available: \url{https://doi.org/10.1145/3638531}
\BIBentrySTDinterwordspacing

\bibitem{PLOTKIN2014127}
D.~Plotkin, ``Chapter 7 - important roles of data stewards,'' in \emph{Data Stewardship}, D.~Plotkin, Ed.\hskip 1em plus 0.5em minus 0.4em\relax Boston: Morgan Kaufmann, 2014, pp. 127--162.

\bibitem{widadQualityAnomalyDetection2023}
E.~Widad, E.~Saida, and Y.~Gahi, ``Quality {{Anomaly Detection Using Predictive Techniques}}: {{An Extensive Big Data Quality Framework}} for {{Reliable Data Analysis}},'' \emph{IEEE Access}, pp. 103\,306--103\,318, 2023.

\bibitem{heinrich2007measure}
\BIBentryALTinterwordspacing
M.~Kaiser, M.~Klier, and B.~Heinrich, ``How to measure data quality? - a metric-based approach,'' in \emph{International Conference on Interaction Sciences}, 2007. [Online]. Available: \url{https://api.semanticscholar.org/CorpusID:15446057}
\BIBentrySTDinterwordspacing

\bibitem{quaresmini2023data}
C.~Quaresmini and G.~Primiero, ``Data quality dimensions for fair ai,'' 2023.

\bibitem{ehrlingerAutomatedContinuousData2018}
L.~Ehrlinger, B.~Werth, W.~Wo{\ss}, and A.~Stra{\ss}e, ``Automated {{Continuous Data Quality Measurement}} with {{QuaIIe}},'' \emph{International Journal On Advances in Software}, 2018.

\bibitem{li2024empowering}
Y.~Li, K.~Ding, J.~Wang, and K.~Lee, ``Empowering large language models for textual data augmentation,'' \emph{arXiv preprint arXiv:2404.17642}, 2024.

\bibitem{moller2023prompt}
A.~G. M{\o}ller, J.~A. Dalsgaard, A.~Pera, and L.~M. Aiello, ``Is a prompt and a few samples all you need? using gpt-4 for data augmentation in low-resource classification tasks,'' \emph{arXiv preprint arXiv:2304.13861}, 2023.

\bibitem{Yang2024llmgpt}
\BIBentryALTinterwordspacing
J.~Yang, H.~Jin, R.~Tang, X.~Han, Q.~Feng, H.~Jiang, S.~Zhong, B.~Yin, and X.~Hu, ``Harnessing the power of llms in practice: A survey on chatgpt and beyond,'' \emph{ACM Trans. Knowl. Discov. Data}, vol.~18, no.~6, apr 2024. [Online]. Available: \url{https://doi.org/10.1145/3649506}
\BIBentrySTDinterwordspacing

\bibitem{ding2024data}
B.~Ding, C.~Qin, R.~Zhao, T.~Luo, X.~Li, G.~Chen, W.~Xia, J.~Hu, A.~T. Luu, and S.~Joty, ``Data augmentation using llms: Data perspectives, learning paradigms and challenges,'' \emph{arXiv preprint arXiv:2403.02990}, 2024.

\bibitem{wangAccuracyWhatData1996}
R.~Y. Wang and D.~M. Strong, ``Beyond {{Accuracy}}: {{What Data Quality Means}} to {{Data Consumers}},'' \emph{Journal of Management Information Systems}, no.~4, pp. 5--33, 1996.

\bibitem{wand1996anchoring}
Y.~Wand and R.~Y. Wang, ``Anchoring data quality dimensions in ontological foundations,'' \emph{Communications of the ACM}, vol.~39, no.~11, pp. 86--95, 1996.

\bibitem{gongSurveyDatasetQuality2023}
Y.~Gong, G.~Liu, Y.~Xue, R.~Li, and L.~Meng, ``A survey on dataset quality in machine learning,'' \emph{Information and Software Technology}, p. 107268, Oct. 2023.

\bibitem{chenPracticalFrameworkEvaluating2019}
H.~Chen, G.~Cao, J.~Chen, and J.~Ding, ``A {{Practical Framework}} for {{Evaluating}} the {{Quality}} of {{Knowledge Graph}},'' in \emph{Knowledge {{Graph}} and {{Semantic Computing}}: {{Knowledge Computing}} and {{Language Understanding}}}, X.~Zhu, B.~Qin, X.~Zhu, M.~Liu, and L.~Qian, Eds.\hskip 1em plus 0.5em minus 0.4em\relax Singapore: Springer Singapore, 2019, pp. 111--122.

\bibitem{foroniEstimatingExtentEffects2021}
D.~Foroni, M.~Lissandrini, and Y.~Velegrakis, ``Estimating the extent of the effects of {{Data Quality}} through {{Observations}},'' in \emph{2021 {{IEEE}} 37th {{International Conference}} on {{Data Engineering}} ({{ICDE}})}.\hskip 1em plus 0.5em minus 0.4em\relax Chania, Greece: IEEE, Apr. 2021, pp. 1913--1918.

\bibitem{pradhanIdentifyingManagingData2023}
S.~K. Pradhan, H.-M. Heyn, and E.~Knauss, ``Identifying and managing data quality requirements: A design science study in the field of automated driving,'' \emph{Software Quality Journal}, May 2023.

\bibitem{cichyOverviewDataQuality2019}
C.~Cichy and S.~Rass, ``An {{Overview}} of {{Data Quality Frameworks}},'' \emph{IEEE Access}, pp. 24\,634--24\,648, 2019.

\bibitem{batiniDataInformationQuality2016a}
C.~Batini and M.~Scannapieco, \emph{Data and {{Information Quality}}: {{Dimensions}}, {{Principles}} and {{Techniques}}}, ser. Data-{{Centric Systems}} and {{Applications}}.\hskip 1em plus 0.5em minus 0.4em\relax Cham: Springer International Publishing, 2016.

\bibitem{schwabeMETRICframeworkAssessingData2024}
D.~Schwabe, K.~Becker, M.~Seyferth, A.~Kla{\ss}, and T.~Sch{\"a}ffter, ``The {{METRIC-framework}} for assessing data quality for trustworthy {{AI}} in medicine: A systematic review,'' Feb. 2024.

\bibitem{DangNLPDQ2024}
V.~M.~H. Dang and R.~M. Verma, ``Data quality in nlp: Metrics and a comprehensive taxonomy,'' in \emph{Advances in Intelligent Data Analysis XXII}, I.~Miliou, N.~Piatkowski, and P.~Papapetrou, Eds.\hskip 1em plus 0.5em minus 0.4em\relax Cham: Springer Nature Switzerland, 2024, pp. 217--229.

\bibitem{wang2023dataforllm}
Z.~Wang, W.~Zhong, Y.~Wang, Q.~Zhu, F.~Mi, B.~Wang, L.~Shang, X.~Jiang, and Q.~Liu, ``Data management for large language models: A survey,'' \emph{arXiv preprint arXiv:2312.01700}, 2023.

\bibitem{whang2023data}
S.~E. Whang, Y.~Roh, H.~Song, and J.-G. Lee, ``Data collection and quality challenges in deep learning: A data-centric ai perspective,'' \emph{The VLDB Journal}, vol.~32, no.~4, pp. 791--813, 2023.

\bibitem{Stvilia2007}
B.~Stvilia, L.~Gasser, M.~Twidale, and L.~Smith, ``A framework for information quality assessment,'' \emph{JASIST}, vol.~58, pp. 1720--1733, 10 2007.

\bibitem{elouataouiAdvancedBigData2022}
W.~Elouataoui, I.~El~Alaoui, S.~El~Mendili, and Y.~Gahi, ``An {{Advanced Big Data Quality Framework Based}} on {{Weighted Metrics}},'' \emph{Big Data and Cognitive Computing}, no.~4, p. 153, Dec. 2022.

\bibitem{dong2015knowledgebased}
X.~L. Dong, E.~Gabrilovich, K.~Murphy, V.~Dang, W.~Horn, C.~Lugaresi, S.~Sun, and W.~Zhang, ``Knowledge-based trust: Estimating the trustworthiness of web sources,'' 2015.

\bibitem{GIL2007227}
Y.~Gil and D.~Artz, ``Towards content trust of web resources,'' \emph{Journal of Web Semantics}, no.~4, pp. 227--239, 2007.

\bibitem{feeneyImprovingCuratedWebData2014}
K.~C. Feeney, D.~O'Sullivan, W.~Tai, and R.~Brennan, ``{Improving Curated Web-Data Quality with Structured Harvesting and Assessment},'' \emph{International Journal on Semantic Web and Information Systems}, no.~2, pp. 35--62, Apr. 2014.

\bibitem{guo2020recommender}
Q.~Guo, F.~Zhuang, C.~Qin, H.~Zhu, X.~Xie, H.~Xiong, and Q.~He, ``A survey on knowledge graph-based recommender systems,'' \emph{IEEE Transactions on Knowledge and Data Engineering}, vol.~34, no.~8, pp. 3549--3568, 2020.

\bibitem{northcutt2021pervasive}
C.~G. Northcutt, A.~Athalye, and J.~Mueller, ``Pervasive label errors in test sets destabilize machine learning benchmarks,'' \emph{arXiv preprint arXiv:2103.14749}, 2021.

\bibitem{gahiMachineLearningDeep2021}
Y.~Gahi and I.~El~Alaoui, ``Machine {{Learning}} and {{Deep Learning Models}} for {{Big Data Issues}},'' in \emph{Machine {{Intelligence}} and {{Big Data Analytics}} for {{Cybersecurity Applications}}}, Y.~Maleh, M.~Shojafar, M.~Alazab, and Y.~Baddi, Eds.\hskip 1em plus 0.5em minus 0.4em\relax Cham: Springer International Publishing, 2021, pp. 29--49.

\bibitem{elouataouiAutomatedBigData2023}
W.~Elouataoui, S.~El~Mendili, and Y.~Gahi, ``An {{Automated Big Data Quality Anomaly Correction Framework Using Predictive Analysis}},'' \emph{Data}, no.~12, p. 182, Dec. 2023.

\bibitem{NithyaDataWork2021}
\BIBentryALTinterwordspacing
N.~Sambasivan, S.~Kapania, H.~Highfill, D.~Akrong, P.~Paritosh, and L.~M. Aroyo, ``“everyone wants to do the model work, not the data work”: Data cascades in high-stakes ai,'' in \emph{Proceedings of the 2021 CHI Conference on Human Factors in Computing Systems}, ser. CHI '21.\hskip 1em plus 0.5em minus 0.4em\relax New York, NY, USA: Association for Computing Machinery, 2021. [Online]. Available: \url{https://doi.org/10.1145/3411764.3445518}
\BIBentrySTDinterwordspacing

\bibitem{Wang2023labelquality}
X.~Wang, X.~Chi, Y.~Song, and Z.~Yang, ``Active learning with label quality control,'' \emph{PeerJ Computer Science}, vol.~9, p. e1480, 09 2023.

\bibitem{Sheng2008label}
\BIBentryALTinterwordspacing
V.~S. Sheng, F.~Provost, and P.~G. Ipeirotis, ``Get another label? improving data quality and data mining using multiple, noisy labelers,'' in \emph{Proceedings of the 14th ACM SIGKDD International Conference on Knowledge Discovery and Data Mining}, ser. KDD '08.\hskip 1em plus 0.5em minus 0.4em\relax New York, NY, USA: Association for Computing Machinery, 2008, p. 614–622. [Online]. Available: \url{https://doi.org/10.1145/1401890.1401965}
\BIBentrySTDinterwordspacing

\bibitem{caiChallengesDataQuality2015}
L.~Cai and Y.~Zhu, ``The {{Challenges}} of {{Data Quality}} and {{Data Quality Assessment}} in the {{Big Data Era}},'' \emph{Data Science Journal}, no.~0, p.~2, May 2015.

\bibitem{JuddooDataGovernance2018}
\BIBentryALTinterwordspacing
S.~Juddoo, C.~George, P.~Duquenoy, and D.~Windridge, ``Data governance in the health industry: Investigating data quality dimensions within a big data context,'' \emph{Applied System Innovation}, vol.~1, no.~4, 2018. [Online]. Available: \url{https://www.mdpi.com/2571-5577/1/4/43}
\BIBentrySTDinterwordspacing

\bibitem{codellaDataQualityChallenges2018}
J.~Codella, C.~Partovian, H.-Y. Chang, and C.-H. Chen, ``Data quality challenges for person-generated health and wellness data,'' \emph{IBM Journal of Research and Development}, no.~1, pp. 3:1--3:8, Jan. 2018.

\bibitem{liHowTextualQuality2020}
L.~Li, T.-T. Goh, and D.~Jin, ``How textual quality of online reviews affect classification performance: A case of deep learning sentiment analysis,'' \emph{Neural Computing and Applications}, no.~9, pp. 4387--4415, May 2020.

\bibitem{ozonzeAutomatingElectronicHealth2023}
O.~Ozonze, P.~J. Scott, and A.~A. Hopgood, ``Automating {{Electronic Health Record Data Quality Assessment}},'' \emph{Journal of Medical Systems}, no.~1, p.~23, Feb. 2023.

\bibitem{ridzuanThematicReviewData2022}
F.~Ridzuan, W.~M.~N. Wan~Zainon, and M.~Zairul, ``A {{Thematic Review}} on {{Data Quality Challenges}} and {{Dimension}} in the {{Era}} of {{Big Data}},'' in \emph{Proceedings of the 12th {{National Technical Seminar}} on {{Unmanned System Technology}} 2020}, K.~Isa, Z.~Md.~Zain, R.~{Mohd-Mokhtar}, M.~Mat~Noh, Z.~H. Ismail, A.~A. Yusof, A.~F. Mohamad~Ayob, S.~S. Azhar~Ali, and H.~Abdul~Kadir, Eds.\hskip 1em plus 0.5em minus 0.4em\relax {Singapore}: {Springer Singapore}, 2022, pp. 725--737.

\bibitem{zhangUnderstandingDetectingDefects2020}
Y.~Zhang and G.~Koru, ``Understanding and detecting defects in healthcare administration data: {{Toward}} higher data quality to better support healthcare operations and decisions,'' \emph{Journal of the American Medical Informatics Association: JAMIA}, no.~3, pp. 386--395, Mar. 2020.

\bibitem{abiyComparisonElectronicMedical2018}
R.~Abiy, K.~Gashu, T.~Asemaw, M.~Mitiku, B.~Fekadie, Z.~Abebaw, A.~Mamuye, A.~Tazebew, A.~Teklu, F.~Nurhussien, M.~Kebede, F.~Fritz, and B.~Tilahun, ``A {{Comparison}} of {{Electronic Medical Record Data}} to {{Paper Records}} in {{Antiretroviral Therapy Clinic}} in {{Ethiopia}}: {{What}} is affecting the {{Quality}} of the {{Data}}?'' \emph{Online Journal of Public Health Informatics}, no.~2, p. e212, 2018.

\bibitem{PansaraCultivatingDataQuality2023}
\BIBentryALTinterwordspacing
R.~Pansara, ``Cultivating data quality to strategies, challenges, and impact on decision-making,'' \emph{International Journal of Managment Education for Sustainable Development}, vol.~6, no.~6, pp. 24--33, 2023. [Online]. Available: \url{https://ijsdcs.com/index.php/IJMESD/article/view/356}
\BIBentrySTDinterwordspacing

\bibitem{mutheeImpactRoutineData2018}
V.~Muthee, A.~F. Bochner, A.~Osterman, N.~Liku, W.~Akhwale, J.~Kwach, M.~Prachi, J.~Wamicwe, J.~Odhiambo, F.~Onyango, and N.~Puttkammer, ``The impact of routine data quality assessments on electronic medical record data quality in {{Kenya}},'' \emph{PloS One}, no.~4, p. e0195362, 2018.

\bibitem{vandenbergheDataQualityAssessment2017}
S.~Van Den~Berghe and K.~Van~Gaeveren, ``Data {{Quality Assessment}} and {{Improvement}}: {{A Vrije Universiteit Brussel Case Study}},'' \emph{Procedia Computer Science}, pp. 32--38, 2017.

\bibitem{kleindienstDataQualityImprovement2017}
D.~Kleindienst, ``The data quality improvement plan: Deciding on choice and sequence of data quality improvements,'' \emph{Electronic Markets}, no.~4, pp. 387--398, Nov. 2017.

\bibitem{abedjanDataProfiling2019}
Z.~Abedjan, L.~Golab, F.~Naumann, and T.~Papenbrock, \emph{Data {{Profiling}}}, ser. Synthesis {{Lectures}} on {{Data Management}}.\hskip 1em plus 0.5em minus 0.4em\relax Cham: Springer International Publishing, 2019.

\bibitem{buelvasDQMANToolMultidimensional2023}
J.~H. Buelvas, D.~M{\'u}nera, and N.~Gaviria, ``{{DQ-MAN}}: {{A}} tool for multi-dimensional data quality analysis in {{IoT-based}} air quality monitoring systems,'' \emph{Internet of Things}, p. 100769, Jul. 2023.

\bibitem{mangDQAguiGraphicalUser2022}
J.~M. Mang, S.~A. Seuchter, C.~Gulden, S.~Schild, D.~Kraska, H.-U. Prokosch, and L.~A. Kapsner, ``{{DQAgui}}: A graphical user interface for the {{MIRACUM}} data quality assessment tool,'' \emph{BMC Medical Informatics and Decision Making}, no.~1, p. 213, Aug. 2022.

\bibitem{whangDataCollectionQuality2023a}
S.~E. Whang, Y.~Roh, H.~Song, and J.-G. Lee, ``Data collection and quality challenges in deep learning: A data-centric {{AI}} perspective,'' \emph{The VLDB Journal}, no.~4, pp. 791--813, Jul. 2023.

\bibitem{hanlonEffectiveUserInterface2021}
R.~Hanlon, M.~Barry, F.~Marrinan, and D.~O'Sullivan, ``Towards an effective user interface for data exploration, data quality assessment and data integration,'' in \emph{2021 {{IEEE}} 15th {{International Conference}} on {{Semantic Computing}} ({{ICSC}})}.\hskip 1em plus 0.5em minus 0.4em\relax Laguna Hills, CA, USA: IEEE, Jan. 2021, pp. 431--436.

\bibitem{Wachirapusitan_2023}
\BIBentryALTinterwordspacing
V.~Wachirapusitan and on~behalf~of CMS~collaboration, ``Machine learning applications for data quality monitoring and data certification within cms,'' \emph{Journal of Physics: Conference Series}, vol. 2438, no.~1, p. 012098, feb 2023. [Online]. Available: \url{https://dx.doi.org/10.1088/1742-6596/2438/1/012098}
\BIBentrySTDinterwordspacing

\bibitem{DolatshahCrowdsourced2018}
\BIBentryALTinterwordspacing
M.~Dolatshah, M.~Teoh, J.~Wang, and J.~Pei, ``Cleaning crowdsourced labels using oracles for statistical classification,'' \emph{Proc. VLDB Endow.}, vol.~12, no.~4, p. 376–389, dec 2018. [Online]. Available: \url{https://doi.org/10.14778/3297753.3297758}
\BIBentrySTDinterwordspacing

\bibitem{LWAKATARE2020}
L.~E. Lwakatare, A.~Raj, I.~Crnkovic, J.~Bosch, and H.~H. Olsson, ``Large-scale machine learning systems in real-world industrial settings: {{A}} review of challenges and solutions,'' \emph{Information and Software Technology}, p. 106368, 2020.

\bibitem{muhammadImprovingLargeData2023}
R.~Y. Muhammad and M.~M. Hamad, ``Improving large data quality through machine learning and artificial intelligence techniques: {{A}} review,'' in \emph{{{3rd International Conference on Smart Cities and Sustainable Planning}}}, Baghdad, Iraq, 2023, p. 020003.

\bibitem{dhoopatiEnhancingEnterpriseApplication2023}
P.~K. Dhoopati, ``Enhancing {{Enterprise Application Integration}} through {{Artificial Intelligence}} and {{Machine Learning}},'' \emph{International Journal of Computer Trends and Technology}, no.~02, pp. 54--60, Feb. 2023.

\bibitem{MAHARANADA202291}
K.~Maharana, S.~Mondal, and B.~Nemade, ``A review: {{Data}} pre-processing and data augmentation techniques,'' \emph{Global Transitions Proceedings}, no.~1, pp. 91--99, 2022.

\bibitem{ChenEmpiricalDA2023}
J.~Chen, D.~Tam, C.~Raffel, M.~Bansal, and D.~Yang, ``An empirical survey of data augmentation for limited data learning in {{NLP}},'' \emph{Transactions of the Association for Computational Linguistics}, pp. 191--211, Mar. 2023.

\bibitem{LIDataAugmentation202271}
\BIBentryALTinterwordspacing
B.~Li, Y.~Hou, and W.~Che, ``Data augmentation approaches in natural language processing: A survey,'' \emph{AI Open}, vol.~3, pp. 71--90, 2022. [Online]. Available: \url{https://www.sciencedirect.com/science/article/pii/S2666651022000080}
\BIBentrySTDinterwordspacing

\bibitem{YangGDA2020}
\BIBentryALTinterwordspacing
Y.~Yang, C.~Malaviya, J.~Fernandez, S.~Swayamdipta, R.~Le~Bras, J.-P. Wang, C.~Bhagavatula, Y.~Choi, and D.~Downey, ``Generative data augmentation for commonsense reasoning,'' in \emph{Findings of the Association for Computational Linguistics: EMNLP 2020}.\hskip 1em plus 0.5em minus 0.4em\relax Association for Computational Linguistics, 2020. [Online]. Available: \url{http://dx.doi.org/10.18653/v1/2020.findings-emnlp.90}
\BIBentrySTDinterwordspacing

\bibitem{dai2023auggpt}
H.~Dai, Z.~Liu, W.~Liao, X.~Huang, Y.~Cao, Z.~Wu, L.~Zhao, S.~Xu, W.~Liu, N.~Liu, S.~Li, D.~Zhu, H.~Cai, L.~Sun, Q.~Li, D.~Shen, T.~Liu, and X.~Li, ``Auggpt: Leveraging chatgpt for text data augmentation,'' 2023.

\bibitem{SUNDBERG2023}
L.~Sundberg and J.~Holmstr{\"o}m, ``Democratizing artificial intelligence: {{How}} no-code {{AI}} can leverage machine learning operations,'' \emph{Business Horizons}, no.~6, pp. 777--788, 2023.

\bibitem{ChengToolSupport2023}
\BIBentryALTinterwordspacing
K.~S. Cheng, P.-C. Huang, T.-H. Ahn, and M.~Song, ``Tool support for improving software quality in machine learning programs,'' \emph{Information}, vol.~14, no.~1, 2023. [Online]. Available: \url{https://www.mdpi.com/2078-2489/14/1/53}
\BIBentrySTDinterwordspacing

\end{thebibliography}

\end{document}